\documentclass[10pt,twocolumn,letterpaper]{article}

\usepackage{cvpr}              

\usepackage{graphicx}
\usepackage{amsmath}
\usepackage{amssymb}
\usepackage{booktabs}

\usepackage{multirow}
\usepackage{color,xcolor}
\usepackage{makecell}
\usepackage{bbding}
\usepackage{soul}

\usepackage{axessibility}

\usepackage[pagebackref,breaklinks,colorlinks]{hyperref}

\usepackage[capitalize]{cleveref}
\crefname{section}{Sec.}{Secs.}
\Crefname{section}{Section}{Sections}
\Crefname{table}{Table}{Tables}
\crefname{table}{Tab.}{Tabs.}

\def\cvprPaperID{3402} 
\def\confName{CVPR}
\def\confYear{2022}

\makeatletter
\def\thanks#1{\protected@xdef\@thanks{\@thanks
        \protect\footnotetext{#1}}}
\makeatother

\begin{document}

\title{Unimodal-Concentrated Loss: Fully Adaptive Label Distribution Learning for Ordinal Regression}

\author{Qiang Li$^*$, Jingjing Wang$^*$, Zhaoliang Yao, Yachun Li, \\
Pengju Yang, Jingwei Yan, Chunmao Wang, Shiliang Pu$^\dag$ \\
\centerline{Hikvision Research Institute, China}\\
\thanks{$^*$Authors contribute equally to this work.}
\thanks{$^\dag$Shiliang Pu is the corresponding author.}
{\tt\small \{liqiang23,wangjingjing9,yaozhaoliang,liyachun6,yangpengju,} \\
{\tt\small yanjingwei,wangchunmao,pushiliang.hri\}@hikvision.com}
}
\maketitle

\begin{abstract}
	Learning from a label distribution has achieved promising results on ordinal regression tasks such as facial age and head pose estimation wherein, the concept of adaptive label distribution learning (ALDL) has drawn lots of attention recently for its superiority in theory. However, compared with the methods assuming fixed form label distribution, ALDL methods have not achieved better performance. We argue that existing ALDL algorithms do not fully exploit the intrinsic properties of ordinal regression. In this paper, we emphatically summarize that learning an adaptive label distribution on ordinal regression tasks should follow three principles. First, the probability corresponding to the ground-truth should be the highest in label distribution. Second, the probabilities of neighboring labels should decrease with the increase of distance away from the ground-truth, i.e., the distribution is unimodal. Third, the label distribution should vary with samples changing, and even be distinct for different instances with the same label, due to the different levels of difficulty and ambiguity. Under the premise of these principles, we propose a novel loss function for fully adaptive label distribution learning, namely unimodal-concentrated loss. Specifically, the unimodal loss derived from the learning to rank strategy constrains the distribution to be unimodal. Furthermore, the estimation error and the variance of the predicted distribution for a specific sample are integrated into the proposed concentrated loss to make the predicted distribution maximize at the ground-truth and vary according to the predicting uncertainty. Extensive experimental results on typical ordinal regression tasks including age and head pose estimation, show the superiority of our proposed unimodal-concentrated loss compared with existing loss functions.
\end{abstract}

\section{Introduction}
\label{sec:intro}

Ordinal regression solves the challenging problems that labels are related in a natural or implied order. Many critical tasks are involved in the ordinal regression problem, e.g., facial age estimation, head pose estimation, facial attractiveness computation and movie ratings, which play an important role in many practical applications such as human-computer interaction, driver monitoring, precise advertising and video surveillance \cite{ 2010Survey, 2012Learning}.

	\begin{figure}[t]
	\begin{center}
		\includegraphics[width=0.9\linewidth]{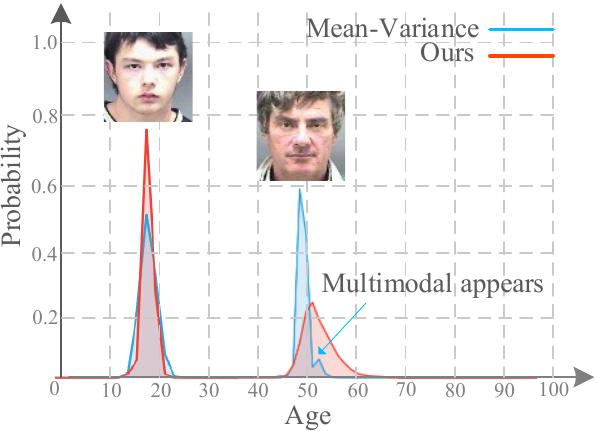}
	\end{center}
	\caption{
		Distributions predicted by Mean-Variance method \cite{2018Mean} and ours.
		Our predictions are optimized to be unimodal and learned according to specific instances adaptively.	
		On the contrary, predictions of Mean-Variance
		are optimized to be  concentrated for all instances
		and do not ensure unimodal distributions explicitly.
	}
	\label{fig:facial_dsitribution}
\end{figure}
	
Early classic works \cite{2004Comparing,2007Demographic,2016Apparent,2015Age,2008Image,2017An} are based on ordinary classification or regression, which do not perform well due to ignoring the ordinal relationship among labels, and suffering from the ambiguous labeling.
In recent years, ranking based methods \cite{ 2016Ordinal,2017Using} are proposed which use multiple binary classifiers to determine the rank order. They explicitly make use of the ordinal information but they do not consider the label ambiguity.

To address the ordinal relationship and label ambiguity, label distribution learning (LDL) \cite{2016LDL} converts a single label to a label distribution. The label distribution covers a certain number of class labels, representing the degree to which each label describes the instance.
Since the real distribution for each instance is not available and must be artificially generated with proper assumption, it can be called fixed form label distribution learning (FLDL). The typical form is the Gaussian distribution centered at the ground-truth with assumed standard deviation \cite{2016LDL,2013TPAMIGeng,2018DLDLv2}. Although FLDL approaches achieve improved performance, however, they use a fixed form distribution to describe various instances which limits their expression ability.

To overcome this limitation, the concept adaptive label distribution learning (ALDL) \cite{2014adaptiveLDL} has been proposed. Among the ALDL based methods, Mean-Variance \cite{2018Mean} is a typical work achieving the promising result, which estimates a distribution with learned mean and variance. However, it pursues a highly concentrated distribution for all instances by making the mean as close to the ground-truth as possible, and the variance as small as possible. Moreover, it can not guarantee the learned distribution is unimodal by a joint use of softmax and mean-variance loss without unimodal constraint. Therefore, we observe that the distributions learned by Mean-Variance are not fully adaptive and are multimodal for some instances, as shown in Fig. \ref{fig:facial_dsitribution}. We can see the learned distribution for the older man is multimodal, and the learned distributions for the two persons are similar. The learned distributions do not accord with the tendency of facial aging, which might be significantly different at different ages \cite{2014adaptiveLDL}.

Obviously, current ALDL methods have not fully exploited the intrinsic properties of ordinal regression. In this paper, the following three principles are summarized for ordinal regression. First, following the empirical risk minimization, the probability corresponding to the ground-truth should be the highest in a label distribution. Second, the labels in ordinal regression tasks change gradually, and the similarity between the test instance and the class prototype decreases gradually when the label move away from the ground-truth. Therefore, the probabilities of neighboring labels accounting for the instance should decrease with the increase of distance away from the ground-truth, i.e., the distribution is unimodal. Third, the label distribution should vary with the samples changing, and even be distinct for different instances with the same
label, due to the different levels of difficulty and ambiguity. In other words, the learned label distribution should be adaptive for a particular instance.
To satisfy the principles above, we propose a new adaptive label distribution learning approach equipped with a unimodal-concentrated loss.
Based on principle \uppercase\expandafter{\romannumeral1}, we directly maximize the probability at the ground-truth via concentrated loss as our primary learning objective. Based on principle \uppercase\expandafter{\romannumeral2}, the unimodal loss derived from learning to rank strategy (LTR) \cite{2016A} is introduced to constrain the distribution to be unimodal. If two neighboring labels are ranked incorrectly, a positive loss would be output to update the trainable parameters to correct the ordinal relationship. Based on principle \uppercase\expandafter{\romannumeral3}, the variance of the distribution corresponding to the concentration degree is integrated and optimized jointly in the concentrated loss, which can be regarded as an indicator of data uncertainty and label ambiguity. The main contributions of this work are three-fold:

\begin{itemize}
	\item
	We are the first to comprehensively summarize the intrinsic principles for learning an adaptive label distribution on ordinal regression tasks. First, the probability at the ground-truth should be the highest in the distribution. Second, the distribution should be unimodal. Third, the distribution should be adaptive to individual instances. These three principles would shed light on the design of loss functions for future works in the field of ordinal regression.
	\item
    Different from previous methods which do not fully comply the above principles, we propose a new unimodal-concentrated loss, with the unimodal part constraining the distribution to be unimodal, and with the concentrated part making the distribution concentrated at the ground-truth and fully adaptive to individual instances.
	\item
	The proposed loss can be easily embedded into existing CNNs without modifying the structure, and extensive experimental results demonstrate its superiority.
\end{itemize}

\section{Related Work}
Existing methods for ordinal regression can be divided into three categories: non-LDL based
methods, FLDL based methods and ALDL based methods.

\subsection{Non-LDL} Non-LDL methods can be grouped into regression based, classification based and ranking based.
Classification based methods usually cast ordinal regression as a classification problem. For examples, age estimation was cast as a classification problem with 101 categories \cite{2016DEX}, and the angle of yaw was divided into coarse bins as class labels for head pose estimation \cite{2014Head,2019Estimation}.
These methods treat ordinal labels as independent ones, and the cost of being assigned to any wrong category is the same, which can’t exploit the relations between labels.
Regression based methods directly regress the ground-truth with Euclidean loss to penalize the difference between the estimation and ground-truth mostly, which do not explicitly make use of the ordinal information. Yi et al. \cite{2015Age} used CNNs models to extract features from several facial regions, and used a square loss for age estimation. Ranjan et al. \cite{2017An} proposed a unified CNN network to jointly estimate facial age, head pose, and other attributes.
Recently, ranking techniques are introduced to the problem of ordinal regression.
Niu et al. \cite{2016Ordinal} leveraged the ordinal information of ages by learning a network with multiple binary outputs, while Chen et al. \cite{2017Using} did this by learning multiple binary CNNs and aggregating the outputs for age estimation.
However, although these methods use ordinal information for better performance, they take a single label as ground-truth without considering label ambiguity.

\subsection{FLDL}
Label distribution learning is proposed to address the label ambiguity issues. For FLDL based methods, distribution form is established before training and kept fixed during training. Their objective is to narrow the gap between the learned distribution and the fixed one.
Geng et al. \cite{2013TPAMIGeng} firstly defined the label distribution by assigning a Gaussian or Triangle distribution for an instance.
DLDL \cite{2017DLDL} adopted the normal distribution and learned the label distribution by minimizing a Kullback-Leibler divergence between two distributions using deep CNNs.
Similar to DLDL, Liu et al. \cite{2020Facial} employed three Gaussian label distributions to describe a face example in the yaw, pitch and roll domain respectively.
DLDL-v2 \cite{2018DLDLv2} improved the DLDL by introducing an expectation loss from distribution to
alleviate the inconsistency between the training objectives and evaluation metric.
DFRs \cite{2017DFR} connected random forests to deep neural networks and
exploited the decision trees’ potential to model any general form of label distributions.
SP-DFRs \cite{2020SPDFR} proposed self-paced regression forests to distinguish noisy and confusing facial images from regular ones, which alleviate the interference arising from them.
However, these methods use a fixed form
distribution to describe various instances which limits their expression ability.

\subsection{ALDL}
Different from FLDL based methods which assume fixed form label distributions, the distribution form for ALDL based methods is not assumed at the beginning and it is generated automatically during learning.
Geng et al. \cite{2013TPAMIGeng} proposed two adaptive label distributions learning algorithms named IIS-ALDL and BFGS-ALDL respectively to automatically learn the label distributions adapted to different ages.
He et al. \cite{2017Data} generated age label distributions through a weighted linear combination of the input image’s label and its context-neighboring samples.
Pan at al. \cite{2018Mean} proposed the Mean-Variance loss, in which the mean loss penalizes the difference between the mean of the estimated distribution and the ground-truth, while the variance loss penalizes the variance of the estimated distribution to ensure a sharp distribution. However, we argue that existing ALDL methods have not strictly complied the intrinsic principles summarized in this work, which can not fully take the advantages of ALDL.

\section{Methodology}

\begin{figure*}[t]
	\begin{center}
		\includegraphics[width=0.85\linewidth]{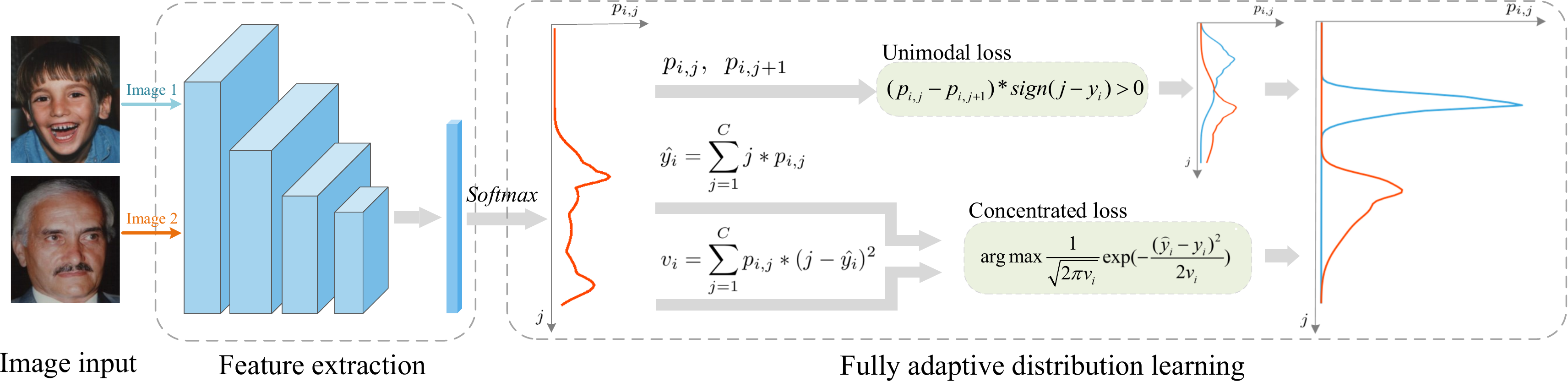}
	\end{center}
	\caption{
		Overview of our proposed method. The unimodal loss makes the final predicted distribution be inclined to a mountain-liking curve with single peak, while the mean and variance of the probabilities are optimized jointly via the concentrated loss to make the predicted distribution adaptive to individual instances.
	}
	\label{fig:method_overview}
\end{figure*}
In this section, we will first give a brief review of FLDL based methods and then detail our ALDL method, where a novel objective function, unimodal-concentrated loss, is proposed for highly flexible distribution learning.

\subsection{Preliminaries}
\label{sec:LDLReview}

Formally, let $x_{i}$ denote the $i$-th input instance with $i=1,2,...,N$, $\hat{y_{i}}$ denote the predicted value by the network, and $y_{i} \in \{1, 2, ..., C\}$ denote the ground-truth label where $N$ is the number of instances and $C$ is the number of classes.
Instead of regressing $y_{i}$ directly, FLDL based methods transform $y_{i}$ from a single class label to
a label distribution and then predict $\hat{y_{i}}$ by label distribution learning.
Gaussian distribution is commonly used in FLDL \cite{2014adaptiveLDL,2016LDL,2017DLDL,2018DLDLv2}.
Instances with the same class label $y_{i}$ share the identical Gaussian distribution.
Taking Gaussian distribution \textbf{\emph{d}}$\sim N(\mu, \sigma^{2})$ as an example
\begin {equation}
d_{i,j} = \frac{1}{S\sqrt{2\pi \sigma^{2}}} \text{exp}( - \frac{(j-\mu)^{2}}{2\sigma^{2}} ), j = 1, 2, ..., C,
\label{eq:dkgt}
\end {equation}
where $d_{i,j}$ denotes the probability of $x_{i}$ belongs to class $j$ and $\sum^{C}_{j}{d_{i,j}}=1$;
$\mu$ equals to the ground-truth label $y_{i}$;
$\sigma$ is the standard deviation of \textbf{\emph{d}}$_{i}$;
$S$ is  a normalization factor.

Let \textbf{\emph{z}}$_{i}=f(x_{i};\Theta)$ denote the output of the last fully connected (FC) layer of a CNN model $f(\cdot)$, where $\Theta$ is the model parameter.
Softmax operation is applied to turn output \textbf{\emph{z}}$_{i}$ into distribution \textbf{\emph{p}}$_{i}$. The elements
$p_{i,j}$ of \textbf{\emph{p}}$_{i}$ is computed as
\begin{equation}
p_{i,j} = \frac{\text{exp}(z_{i,j})}{\sum^{C}_{k=1}\text{exp}(z_{i,k})}.
\label{eq:pik}
\end{equation}
Kullback-Leibler (KL) divergence is usually adopted in FLDL as the loss function.
KL loss $(L_{KL})$ is optimized to reduce the gap between the pre-defined distribution \textbf{\emph{d}}$_{i}$ and the predicted distribution \textbf{\emph{p}}$_{i}$.
The final prediction $\hat{y_{i}}$ is obtained by taking the expectation of \textbf{\emph{p}}$_{i}$ as follows
\begin{equation}
\hat{y_{i}} = \sum^{C}_{j=1}j*p_{i,j}.
\label{eq:haty}
\end{equation}
Thus, different instances with the same label are expected to predict similar distributions. It is against the nature that different instances with the same label should have their own distributions corresponding to their characteristics.

\subsection{Proposed Approach}
\label{sec:overview}
In order to tackle the issues above, we present a novel adaptive label distribution learning method which can produce unimodal and instance-aware distributions. Fig. \ref{fig:method_overview} gives the overview of our approach, in which the proposed unimodel loss and concentrated loss are embedded into a exsting CNN for end-to-end learning without any additional modification on the model. The details are given below.

\subsubsection{Unimodal loss}
Based on the principles we have summarized previously, it is crucial to output a unimodal distribution for ordinal regression tasks.
Hence, we propose a unimodal loss denoted as $L_{uni}$, which is formulated as follows
\begin{equation}
L_{uni} = \frac{1}{N}\sum_{i=1}^{N}
\sum^{C-1}_{j=1}\text{max}(0,{-}(p_{i,j}{-}p_{i,j+1})*\text{sign}[j-y_{i}]),
\label{eq:unimodal_loss_1}
\end{equation}
where $\text{sign}[j-y_{i}]$ is a sign function which equals to -1 while $j-y_{i}<0$ and equals to 1 otherwise. It is desirable for value of $p_{i,j}-p_{i,j+1}$ to be negative if $j-y_{i}<0$ and be positive if $j-y_{i}>0$, which conforms to the properties of unimodal distribution.

\textbf{Constrain distribution to be unimodal}.
In order to show how our unimodal loss $L_{uni}$ performs, we take a case of $j<y_{i}$ (i.e. \text{sign}$[j-y_{i}]{=}-1$) for illustration, as shown in the blue region of Fig. \ref{fig:unimodal_1}, where $p_{i,j}-p_{i,j+1}>0$. That is the adjacent probabilities are not in ascending order, and consequently the gradient of $L_{uni}$ w.r.t. $p_{i,j}$ and $p_{i,j+1}$ can be computed respectively as

\begin{equation}
\frac{\partial L_{uni}}{\partial p_{i,j}} = +1,
\label{eq:gradient_luni_1}
\end{equation}
\begin{equation}
\frac{\partial L_{uni}}{\partial p_{i,j+1}} = -1.
\label{eq:gradient_luni_2}
\end{equation}
According to Eq. \ref{eq:gradient_luni_1} and Eq. \ref{eq:gradient_luni_2}, the $p_{i,j}$ will be decreased due to its positive gradients, while $p_{i,j+1}$ will be increased due to its negative gradients.
In other words, our unimodal loss $L_{uni}$ adjusts the probabilities to make them increase monotonically before reaching the ground-truth position.

In the other direction where $\text{sign}[j-y_{i}]{=}+1$, our $L_{uni}$ adjusts the probabilities to decrease monotonically after the ground-truth position.
Thus, the predicted distribution will be optimized to be unimodal via $L_{uni}$.

Our proposed $L_{uni}$ is superior to the softmax loss used in \cite{2018Mean}.
since $L_{uni}$ can adjust the ranking relation within the predicted distribution while the softmax loss not. Please refer to proof in Sec. \ref{Comparison with Mean-Variance} for more details. Consequently, the predicted probabilities of Mean-Variance \cite{2018Mean} are more likely to be multimodal, and the examples for comparison are given in Fig. \ref{fig:age_figure_new}.

\subsubsection{Concentrated loss}
According to principles discussed before, the learned distribution should maximize at the ground-truth and be adaptive for individual instances.
To accomplish this goal, we propose a concentrated loss denoted as $L_{con}$, which integrates the difference between the estimation $\hat{y}$ and the ground-truth $y$ and the uncertainty indictor variance of the predicted distribution together, and optimizes them jointly.

We first maximize the following likelihood for $x_{i}$
\begin{equation}
\Phi(\textbf{\emph{p}}_{i}; x_{i}, \Theta) = \frac{1}{N} \sum^{N}_{i=1} \frac{1}{\sqrt{2\pi v_{i}}}\text{exp}{(}{-} \frac{ (\hat{y_{i}} - y_{i})^{2}  }{2v_{i}}{)} ,
\label{eq:concentrate_loss_1}
\end{equation}
where $v_{i}$ is the variance of predicted distribution \textbf{\emph{p}}$_{i}$. Based on Eq. \ref{eq:pik} and Eq. \ref{eq:haty}, $v_{i}$ can be calculated as below
\begin{equation}
v_{i} = \sum^{C}_{j=1}p_{i,j}*(j-\hat{y_{i}})^{2}.
\label{eq:vi}
\end{equation}
Then we take the negative log of $\Phi(\cdot)$ to get $L_{con}$ as follows
\begin{align}
L_{con} &= - \text{ln} (\Phi(\textbf{\emph{p}}_{i}; x_{i}, \Theta) )\\ &= \frac{1}{N} \sum^{N}_{i=1} (
\frac{1}{2}\text{ln}v_{i} {+}  \frac{ (\hat{y_{i}}- y_{i})^{2} }{2v_{i}} {+} \frac{1}{2}\text{ln}2\pi),
\label{eq:concentrate_loss_21}
\end{align}
where constant $\frac{1}{2}\text{ln}2\pi$ can be omitted during optimization.

\textbf{Instance-aware adaptive distribution learning}.
\label{sub:instance aware distribution}
To demonstrate how it works,
we take the gradient of concentrated loss $L_{con}$ w.r.t. the variance $v_{i}$. As we all know the sample mean and variance are statistically independent of each other, for simplicity, it is computed as
\begin{equation}
\frac{\partial L_{con}}{\partial v_{i}} = \frac{1}{2v_{i}} - \frac{  (\hat{y_{i}} - y_{i})^{2} }{2v^{2}_{i}},
\label{eq:gradient_lcon_1}
\end{equation}
where $\frac{\partial L_{con}}{\partial v_{i}}$ has following properties
\begin{equation}
\frac{\partial L_{con}}{\partial v_{i}} > 0, \ \text{while} \ v_{i}> (\hat{y_{i}} - y_{i})^{2}, \quad \ \ \
\label{eq:gradient_lcon_2}
\end{equation}
\begin{equation}
\frac{\partial L_{con}}{\partial v_{i}} < 0, \text{while} \ 0<v_{i}< (\hat{y_{i}} - y_{i})^{2}. \
\label{eq:gradient_lcon_3}
\end{equation}

According to Eq. \ref{eq:gradient_lcon_2}, the network will be optimized to decrease the intensity of $v_{i}$ to make it close to $(\hat{y_{i}} - y_{i})^{2}$ via its positive gradient.
In this situation, $(\hat{y_{i}} - y_{i})^{2}$ is an adaptive lower bound of $v_{i}$.
In other words, when estimation error $(\hat{y_{i}} - y_{i})^{2}$ is small which indicates an easy sample, the distribution variance $v_{i}$ is decreased to be small.

According to Eq. \ref{eq:gradient_lcon_3},  the network will be optimized to increase the intensity of $v_{i}$ to make it close to $(\hat{y_{i}} - y_{i})^{2}$ via its negative gradient.
In this situation, $ (\hat{y_{i}}- y_{i})^{2}$ is an adaptive upper bound of $v_{i}$.
That is to say, when estimation error $(\hat{y_{i}} - y_{i})^{2}$ is large which indicates a hard sample, the distribution variance $v_{i}$ is increased to be large.

Take the gradient of $L_{con}$ w.r.t. the estimation error $\epsilon_{i}= (\hat{y_{i}}- y_{i})^{2}$ as follows
\begin{equation}
\frac{\partial L_{con}}{\partial \epsilon_{i}} = \frac{1}{2v_{i}}.
\label{eq:gradient_lcon_4}
\end{equation}
According to Eq. \ref{eq:gradient_lcon_4},
the estimation error $\epsilon$ is always optimized as small as possible via its positive gradient. Moreover,
the optimization speed of $\epsilon$ is negatively
correlated with the magnitude of $v_{i}$.

Finally, the estimation error and the variance of the distribution are optimized in a fully adaptive way, and consequently the learned distribution can be instance-aware. As shown in Fig. \ref{fig:ages_headpose_std}, the first row examples are in high quality and the second row examples are in low quality which are polluted by illumination, occlusion and heavy makeup. It is obvious that our predicted distributions can reflect the quality among faces where variances of the first row instances are small while the variances of the second ones are large.

\subsubsection{Unimodal-Concentrated loss}
The final objective function of our proposed approach is denoted as $L_{uc}$ and formulated as follows
\begin{equation}
L_{uc} = L_{con} + \lambda * L_{uni},
\label{eq:final_loss}
\end{equation}
where $\lambda$ is a hyper-parameter to weight the two terms.


\begin{figure}[t]
	\begin{center}
		\includegraphics[width=0.8\linewidth]{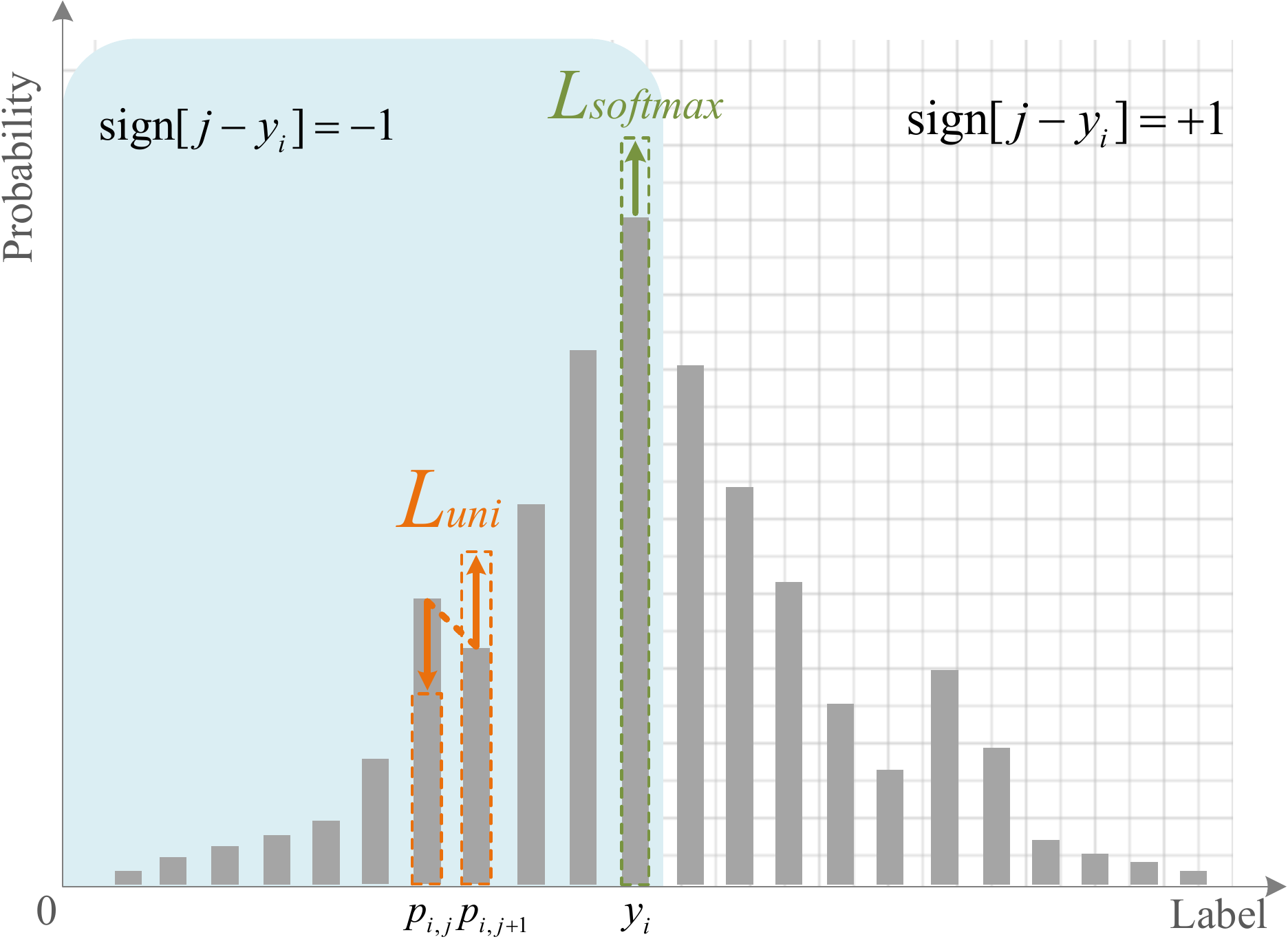}
	\end{center}
	\caption{
		An illustration of how unimodal loss (orange) and softmax loss (green) affect the probability distribution respectively.}
	\label{fig:unimodal_1}
\end{figure}

\textbf{Comparison with Mean-Variance}.
\label{Comparison with Mean-Variance}
The Mean-Variance loss \cite{2018Mean} can be formulated as
\begin{align}
	L_{m-v}&=L_{s} + \lambda_{1}L_{m} + \lambda_{2}L_{v} \\
   	  &=\frac{1}{N}\sum_{i=1}^{N}{-}logp_{i,y_{i}} +\frac{\lambda_{1}}{2}(\hat{y_{i}} - y_{i})^2 + \lambda_{2}v_{i},
	\label{eq:mean-variance}
\end{align}
where $L_{s}$ is the softmax loss. To show the effect of Mean-Variance loss on the generated distribution, we take the gradient of $L_{m-v}$ with respect to item $p_{i,j}$, $(\hat{y_{i}}-y_{i})$ and $v_{i}$, respectively.
Firstly, we take the gradient of $L_{m-v}$ w.r.t $p_{i,j}$
\begin{align}
\frac{\partial L_{m-v}}{\partial p_{i,j}}
=
\left\{
\begin{array}{lr}
\lambda_{1}(\hat{y_{i}}{-}y_{i})j{+}\lambda_{2}(j{-}\hat{y_{i}})^2, \quad \quad \ \ \  j {\neq} y_{i} \\
\frac{-1}{p_{i,j}} + \lambda_{1}(\hat{y_{i}}{-}y_{i})j{+}\lambda_{2}(j{-}\hat{y_{i}})^2,  j{=}y_{i}. \\
\end{array}
\right.
\label{eq:lmv-pij}
\end{align}
And then, we take the gradient of $L_{m-v}$ w.r.t $\epsilon_{i}= (\hat{y_{i}}- y_{i})^{2}$
\begin{equation}
	\frac{\partial L_{m{-}v}}{\partial \epsilon_{i}} = \frac{1}{2}\lambda_{1}.
	\label{eq:lmv-yi}
\end{equation}
Finally, we take the gradient of $L_{m-v}$ w.r.t $v_{i}$
\begin{equation}
	\frac{\partial L_{m{-}v}}{\partial v_{i}} = \lambda_{2}.
	\label{eq:lmv-vi}
\end{equation}
For simplicity, we omit $\frac{1}{N}$ in Eq.\ref{eq:lmv-pij}, Eq.\ref{eq:lmv-yi}, Eq.\ref{eq:lmv-vi}.

Base on equations above, we have three observations:
\begin{itemize}
\item
According to Eq.\ref{eq:lmv-pij}, we can see that the gradient $\frac{\partial L_{m-v}}{\partial p_{i,j}}$ and $\frac{\partial L_{m-v}}{\partial p_{i,j+1}}$ have similar expression and the direction of both $\frac{\partial L_{m-v}}{\partial p_{i,j}}$ and $\frac{\partial L_{m-v}}{\partial p_{i,j+1}}$ are irrelevant with the relative order of $p_{i,j}$ and $p_{i,j+1}$. Additionally, $\frac{\partial L_{s}}{\partial p_{i,j}}$ can only be non-zero when $j={y_{i}}$ which means that the softmax loss cannot correct the wrong ordinal relationship between adjacent probabilities, see Fig. \ref{fig:unimodal_1}.
\item
According to Eq.\ref{eq:lmv-vi}, the gradient $\frac{\partial L_{m{-}v}}{\partial v_{i}}$  is a positive constant which means that Mean-Variance loss will always optimize the variance of the predicted distribution to be small. In other words, Mean-Variance loss makes the estimated distribution as sharp as possible \cite{2018Mean}.
\item
According to Eq.\ref{eq:lmv-yi} and Eq.\ref{eq:lmv-vi}, we can see that there is no $v_i$ item in gradient $\frac{\partial L_{m{-}v}}{\partial (\hat{y_{i}}-y_{i})^2}$ and there is no $(\hat{y_{i}}-y_{i})$ item in gradient $\frac{\partial L_{m{-}v}}{\partial v_{i}}$.
That is to say, the estimation error and variance of the predicted distribution are optimized independently without interaction.
\end{itemize}

In summary, the Mean-Variance loss does not constrain the predicted distribution to be unimodal explicitly.
Besides, the minimization of  Mean-Variance loss does not generate an instance-aware distribution adaptively.


\section{Experiments}
In this section, we will first detail the experiment settings and then compare our method with state-of-the-art works on facial age database MORPH Album II \cite{2006MORPH} and head pose databases including AFLW2000 \cite{2016Face} and BIWI \cite{fanelli2011real}.

\subsection{Datasets}
\textbf{MORPH Album II} is one of the most commonly used and largest longitudianal face databases in the public domain for age estimation, which contains 55,134 face images of 13,617 subjects and the ages range from 16 to 77 \cite{2006MORPH}.
Mugshots are captured in high quality and all faces are frontal. We follow the most widely adopted evaluation protocol namely the five-fold random split (RS) protocol \cite{2017Using,2018Mean,2018DLDLv2,2020SPDFR}, where 80 percent of images are randomly chosen as the training set and the remaining for testing.

\textbf{IMDB-WIKI} contains more than half a million labeled images of celebrities, which are crawled from IMDB and Wikipedia. Although it is the largest facial dataset with age labels, it is polluted by too much noise. Instead of using it to evaluate our method, we utilize it to pre-train our network as previous works \cite{2016Deep, 2018DLDLv2, 2020BridgeNet}.

\begin{table}[b]
	\caption{Comparisons with other state-of-the-art methods on the Morph II. All results are under the five-fold random split protocol.}
	\begin{center}
		\begin{tabular}{|p{4.8cm}<{\centering}|p{1.5cm}<{\centering}|p{0.6cm}<{\centering}|}
			\hline
			Method  & Form & {MAE} \\
			\hline\hline
			Ranking-CNN \cite{2017Using} &Non-LDL& 2.96  \\  \hline
			BridgeNet \cite{2020BridgeNet} &Non-LDL&2.38  \\ \hline
			DLDL-v2 \cite{2018DLDLv2} &FLDL& 1.97 \\ \hline
			DRFs \cite{2017DFR} &FLDL& 2.17 \\  \hline
			SPUDRFs \cite{2020SPDFR} &FLDL& 1.91 \\ \hline
			Mean-Variance \cite{2018Mean} &ALDL& 2.16  \\ \hline
			AVDL \cite{2020Adaptive} &ALDL& 1.94  \\  \hline
			\hline
			Ours &ALDL& \bf{1.86 } \\
			\hline
		\end{tabular}
	\end{center}
	\label{table:age_compare}
\end{table}

\begin{table*}[t]
	\caption{Comparisons with other state-of-the-art methods on AFLW2000 and BIWI dataset. All models are trained on 300W-LP dataset. }
	\begin{center}
		\small
		\setlength{\tabcolsep}{1.8mm}{
			\begin{tabular}{|l|c|c|c|c|c|c|c|c|c|}
				\hline
				\multirow{2}{*}{Method}&\multirow{2}{*}{Form}&\multicolumn{4}{c|}{AFLW2000}&\multicolumn{4}{c|}{BIWI} \cr \cline{3-10}
				&&Yaw&Pitch&Roll&Mean&Yaw&Pitch&Roll&Mean \\
				\hline
				\hline
				3DDFA \cite{2016Face} &Non-LDL&5.40 & 8.53 & 8.25 & 7.39 & 36.17 & 12.25 & 8.77 & 19.06 \\ \hline
				FAN \cite{2017How} &Non-LDL& 6.36 & 12.28 & 8.71 & 9.12 & 8.53 & 7.48 & 7.63 & 7.88 \\ \hline
				Hopenet ($\alpha$=2) \cite{2018Fine} &Non-LDL& 6.47 & 6.56 & 5.44 & 6.16 & 5.17 &6.98 & 3.39 & 5.18 \\ \hline
				Hybrid Classification \cite{2019A} &Non-LDL& 4.82 & 6.23 & 5.14 & 5.40 & - & - & - & - \\ \hline
				FSA \cite{2020FSA} &Non-LDL& 4.50 & 6.08& 4.64 & 5.07 & 4.27 & 4.96 & 2.76 & 4.00 \\ \hline
				FDN \cite{2020FDN} &Non-LDL& 3.78 & 5.61 & 3.88 & 4.42 & 4.52 & 4.70 & \bf{2.56} & 3.93 \\ \hline
				Guo \cite{2021order} &Non-LDL& - & - & - & - & \bf{3.68} & 4.36 & 3.02 & 3.69 \\ \hline
				\hline
				Ours &ALDL& \bf{3.46} & \bf{5.24} & \bf{3.68} & \bf{4.13} & 3.91 & \bf{3.96} & 2.83 & \bf{3.57} \\ \hline
			\end{tabular}
		}
	\end{center}
	\label{tabel:compare_head_pose}
\end{table*}

\begin{table}[t]
	\caption{The performances compared with the same backbone network but different losses. }
	\begin{center}
		\small
		\begin{tabular}{|c|c|c|c|c|}
			\hline
			Loss  & Form & MORPH II & AFLW&BIWI  \\
			\hline\hline
			DLDL-v2 & FLDL & 1.90 & 4.20 & 3.80 \\ \hline
			Mean-Variance &  ALDL  & 2.01 & 4.36 & 4.01 \\ \hline
			\hline
			Ours &  ALDL & \bf{1.86}& \bf{4.13} & \bf{3.57} \\
			\hline
		\end{tabular}
	\end{center}
	\label{table:compare_fixed_adaptive_distribution}
\end{table}

\begin{table}[t]
	\caption{The results for different loss combinations of Mean-Variance and ours.}
	\begin{center}
		\small
		\setlength{\tabcolsep}{1.2mm}{
			\begin{tabular}{|c|c|c|c|c|}
				\hline
				\multicolumn{2}{|c|}{Combinations} & \multicolumn{3}{c|}{Benchmarks} \cr \cline{1-5}
				Auxiliary& Primary & MORPH II & AFLW2000 & BIWI \\ \hline \hline
				Softmax &Concentrated& 1.92 & 4.25 & 3.61 \\ \hline
				Unimodal &Concentrated&1.86 & 4.13 & 3.57 \\ \hline
				Softmax &Mean \& Variance &2.01 & 4.36 & 4.01 \\ \hline
				Unimodal &Mean \& Variance & 3.30 & 4.53 & 4.39 \\ \hline
			\end{tabular}
		}
	\end{center}
	\label{tabel:cross_validation}
\end{table}

\textbf{AFLW2000} is one of the most commonly used benchmarks for head pose estimation \cite{2018Fine, 2020FSA, 2020FDN}. The challenging AFLW2000 dataset \cite{2016Face} contains the first 2,000 samples of the AFLW dataset \cite{M2012Annotated} which have been re-annotated with 68 3D landmarks using a 3D model for each face. The faces in the dataset have large pose variations with various occlusions, expressions as well as illumination conditions.

\textbf{BIWI} is collected by recording RGB-D videos of 20 different subjects across different head poses using a Kinect v2 device in a laboratory setting, and about 15,000 frames are generated with pose annotations \cite{fanelli2011real} .

\textbf{300W-LP} dataset \cite{2016Face} is re-annotated from a collection of several popular in the wild facial 2D landmark datasets by fitting the 3D dense face model to the image. The database contains 61,225 samples across large poses and expands to 122,450 samples by horizontal-flipping.
Following the previous works \cite{2018Fine, 2020FSA, 2020FDN},
we use 300W-LP dataset for network training while using AFLW2000 and BIWI for evaluation.


\subsection{Implementation Details}
\label{implementation}
For the age estimation task, we use VGG-16 \cite{2014vgg16} as the backbone network without modification except the dimension of the last fully-connected layer is modified to 101 for wide age range following \cite{2017DLDL,2018DLDLv2,2018Mean}.
All faces are cropped and resized to the $224\times224$ resolution. Data augmentation includes random horizontal flipping, standard color jittering and random affine transformation.
The model is pre-trained on IMDB-WIKI and then fine-tuned with a learning rate $lr$ which is initialized as $lr{=}0.01$ and decayed by a factor of 0.5 after each 10$K$ iterations. the maximum number of iterations  is 60$K$, and batch size is set to 128.
Hyper-parameter $\lambda$ is 1000.

For the head pose estimation task, we directly follow the experiment settings of Hopenet \cite{2018Fine}, in which Resnet-50 \cite{2016resnet} is chosen as the backbone network and Adam optimizer \cite{2014Adam} is used for optimization.
Please kindly refer to Hopenet for more experiment details if you need.
It's worth to note that, we also make a modification like Hopenet, i.e., the output dimension is changed from 66 to 200 for the reason that the angles are in $\pm99^{\circ}$ in fact.	
In this way, 31 images are discarded from AFLW2000 for their angles are out of range.
Hyper-parameter $\lambda$ is set to be 1000.
Following \cite{2017DLDL,2018Mean, 2018Fine}, we use MAE as our evaluation metric for both tasks.

\begin{figure}[t]
	\begin{center}
		\includegraphics[width=0.8\linewidth]{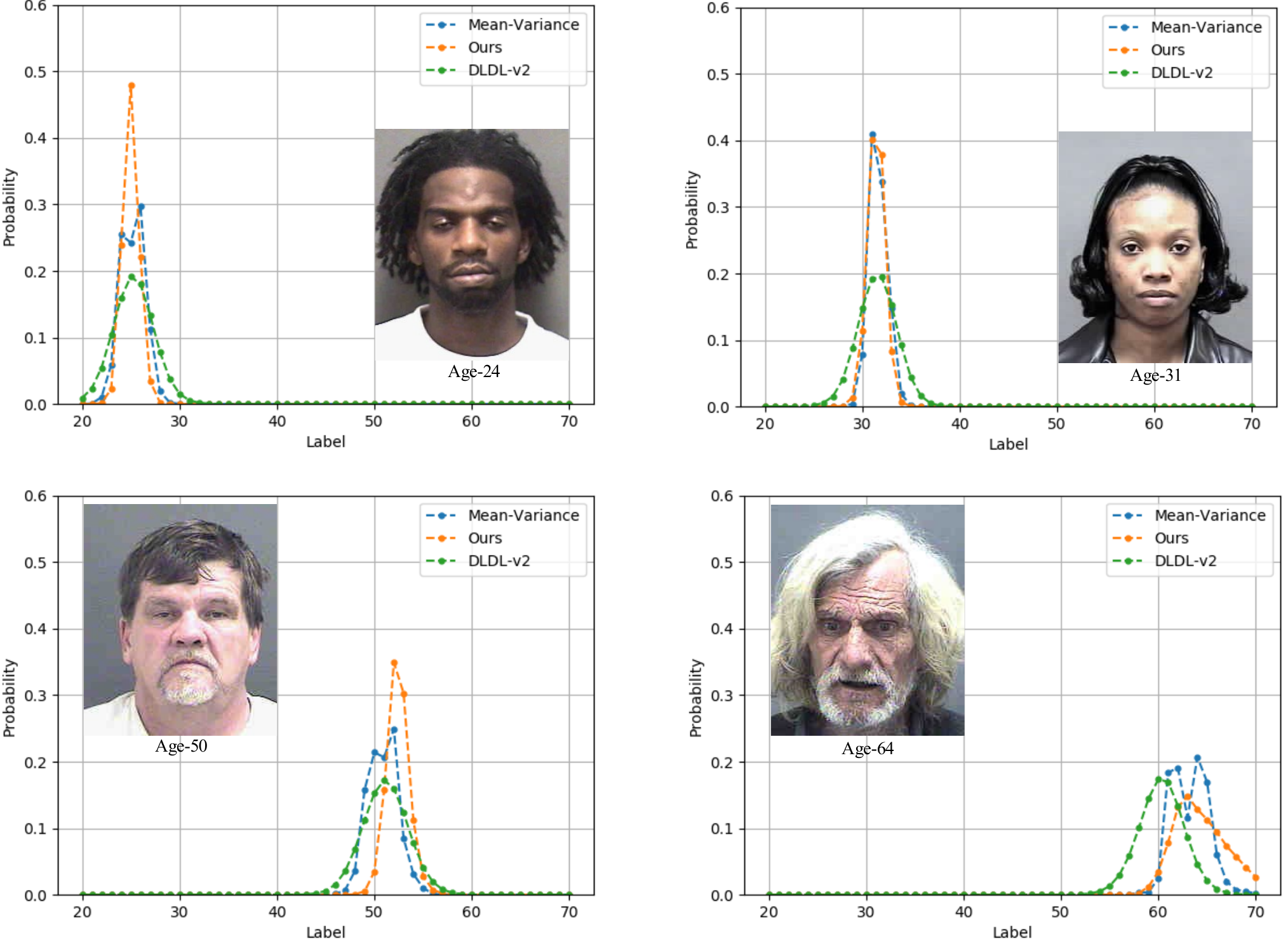}
	\end{center}
	\caption{Age examples for comparisons with the same backbone network but different losses. Some distributions are not unimodal generated by Mean-Variance while our proposed method can ensure the unimodality of distribution. And DLDL-v2 tends to output the distributions with similar shapes.
	}
	\label{fig:age_figure_new}
\end{figure}


\begin{figure*}[t]
	\centering
	\includegraphics[width=0.2 \linewidth]{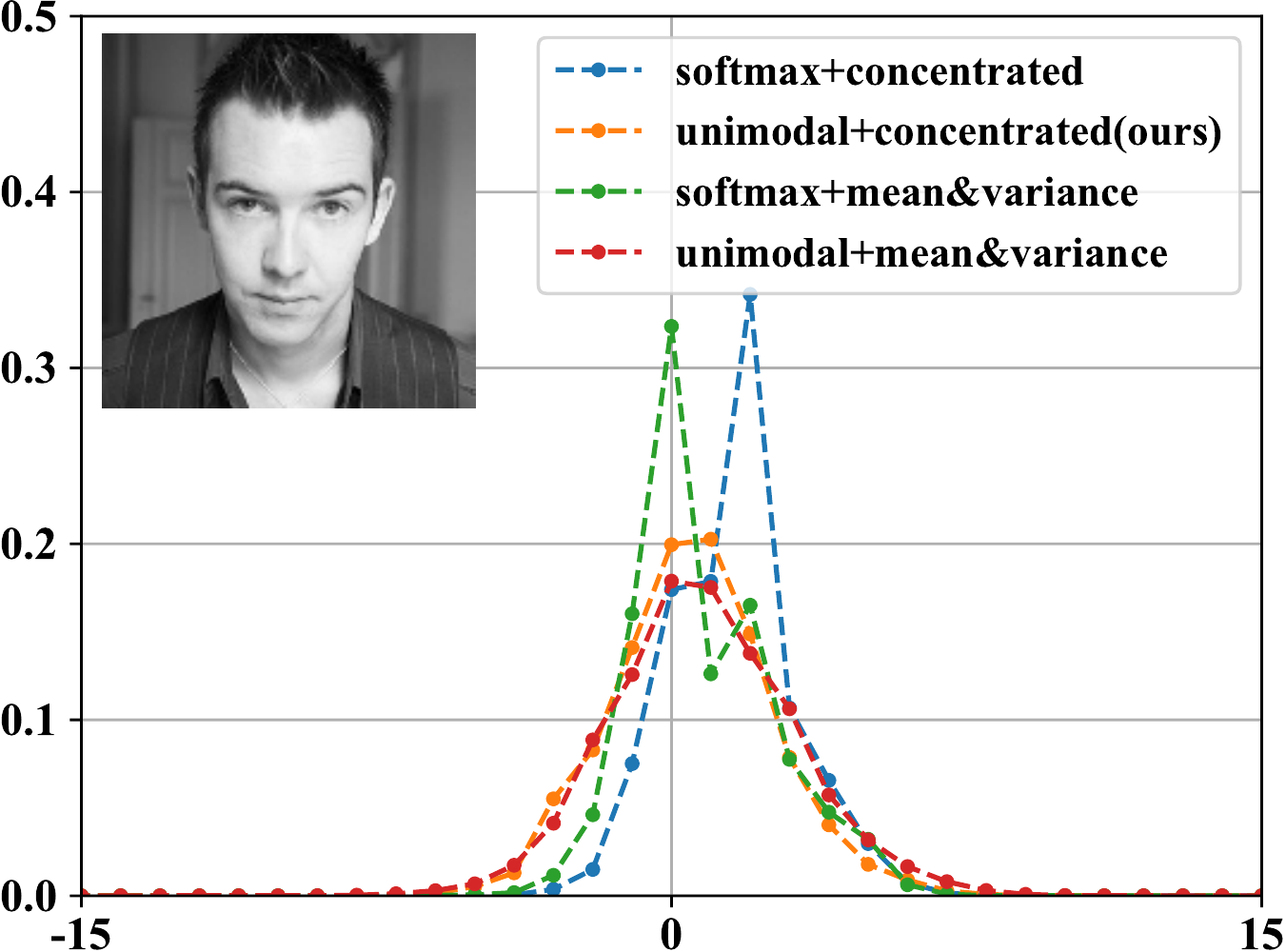} \quad
	\includegraphics[width=0.2 \linewidth]{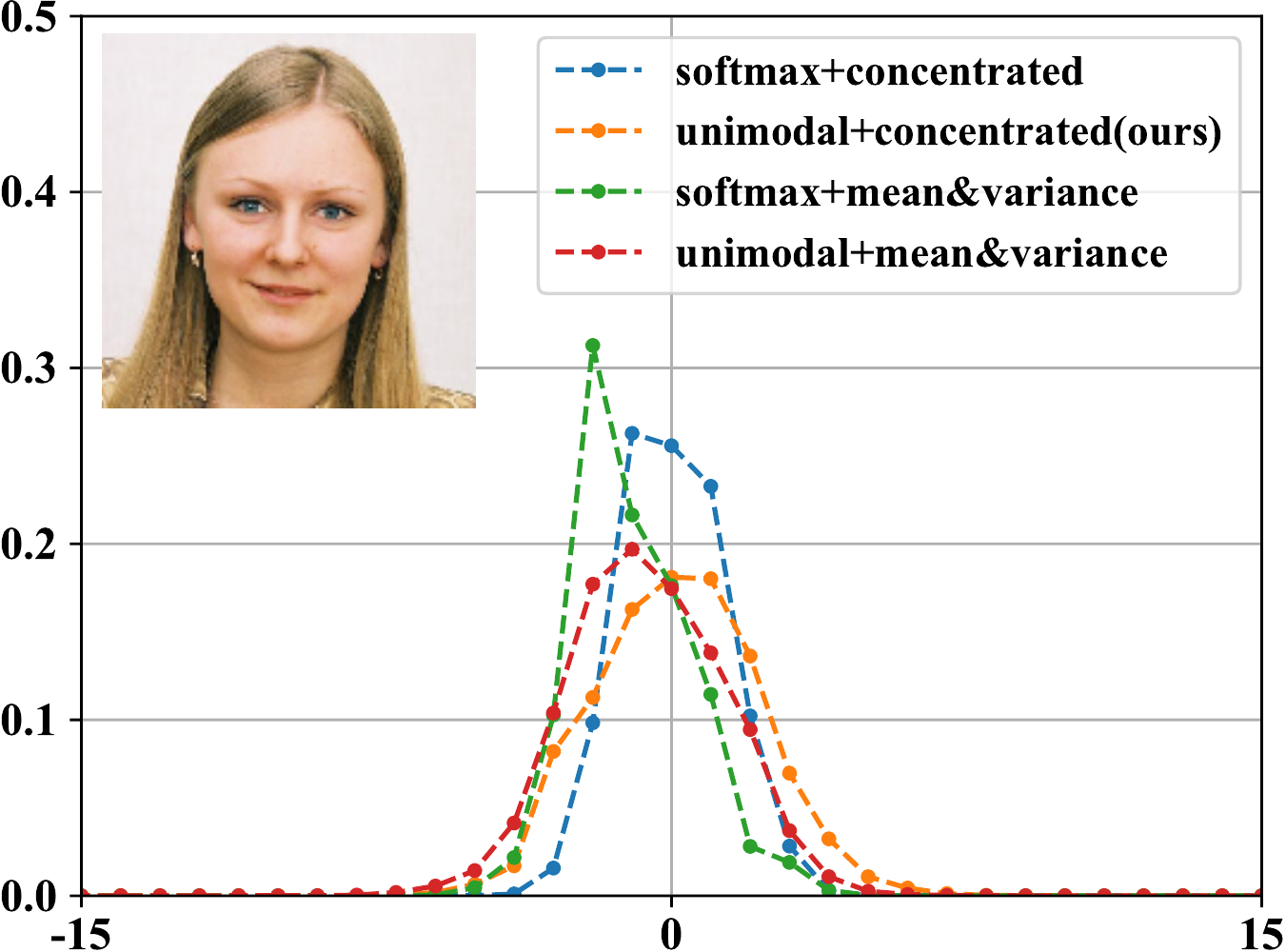} \quad
	\includegraphics[width=0.2 \linewidth]{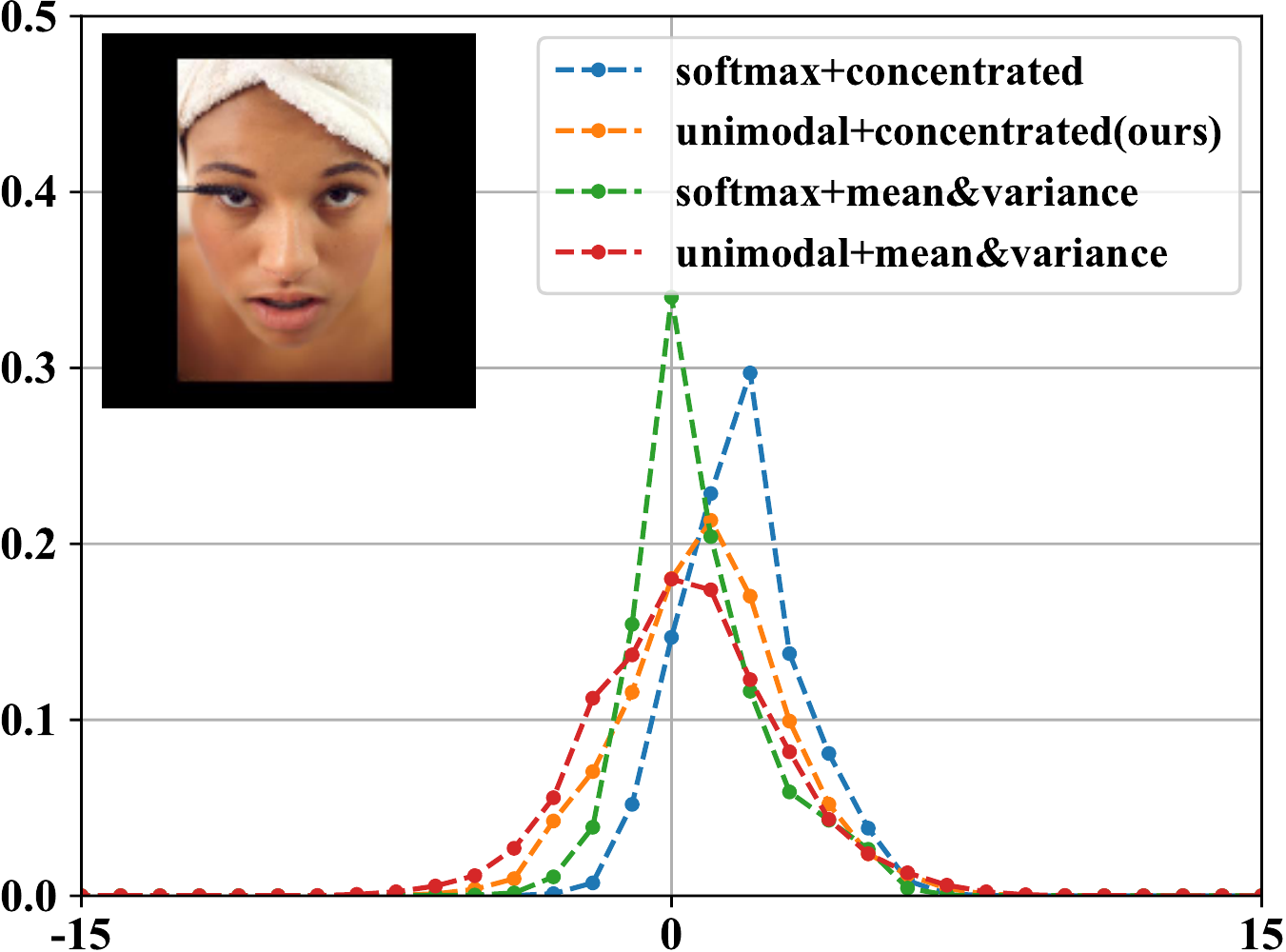} \quad
	\includegraphics[width=0.2 \linewidth]{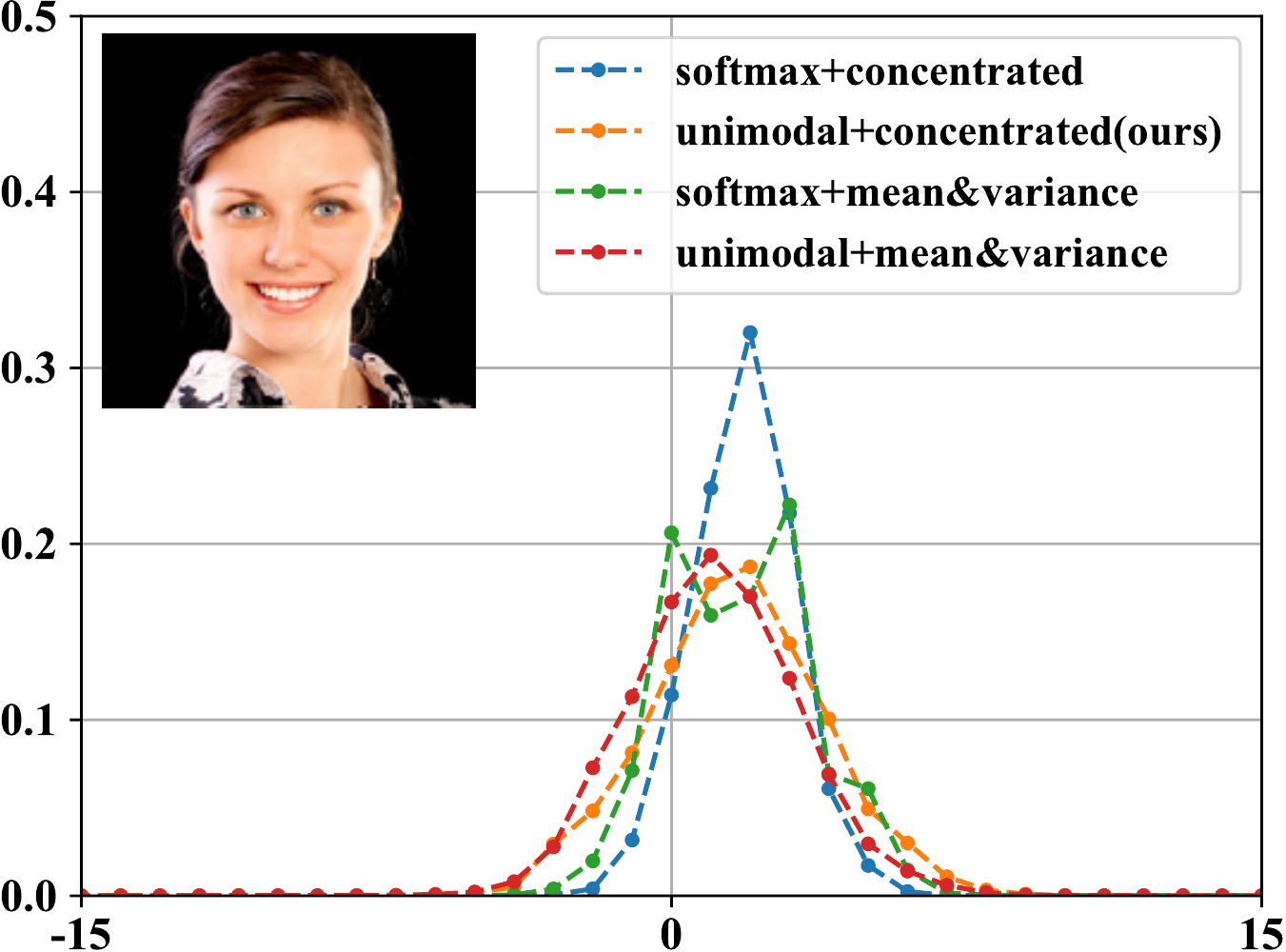}
	\\[1ex] \
	\includegraphics[width=0.2 \linewidth]{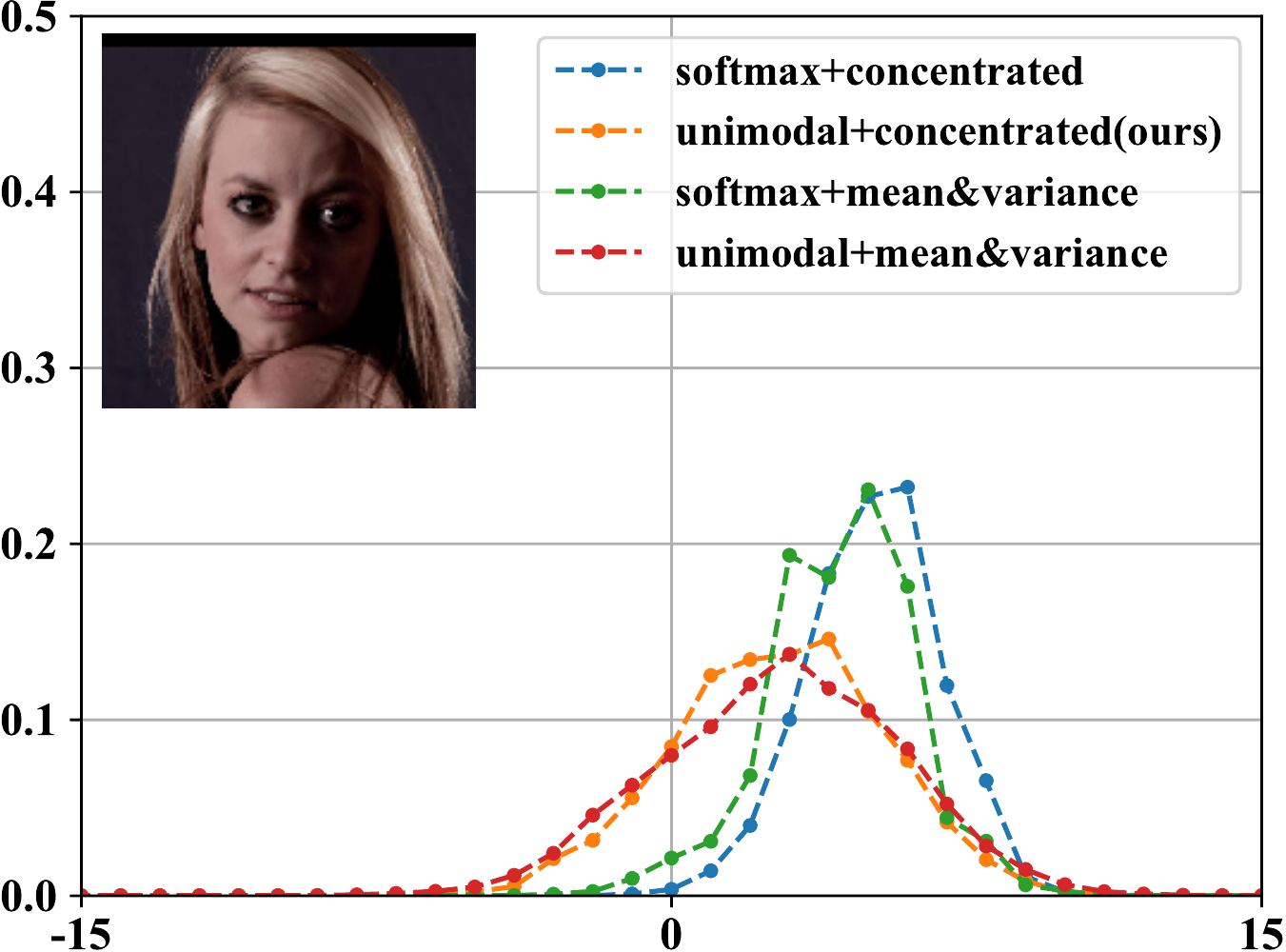} \quad
	\includegraphics[width=0.2 \linewidth]{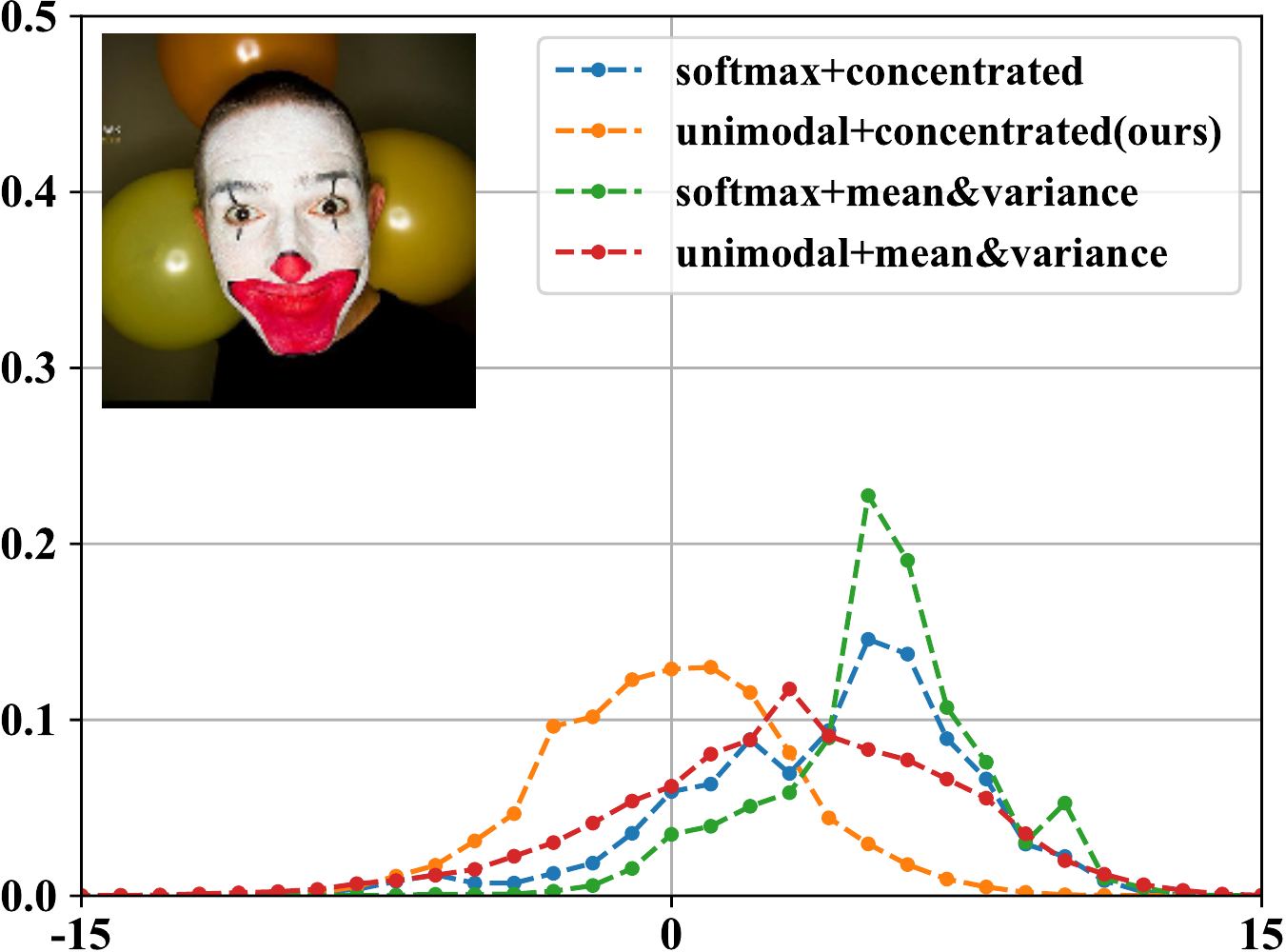} \quad
	\includegraphics[width=0.2 \linewidth]{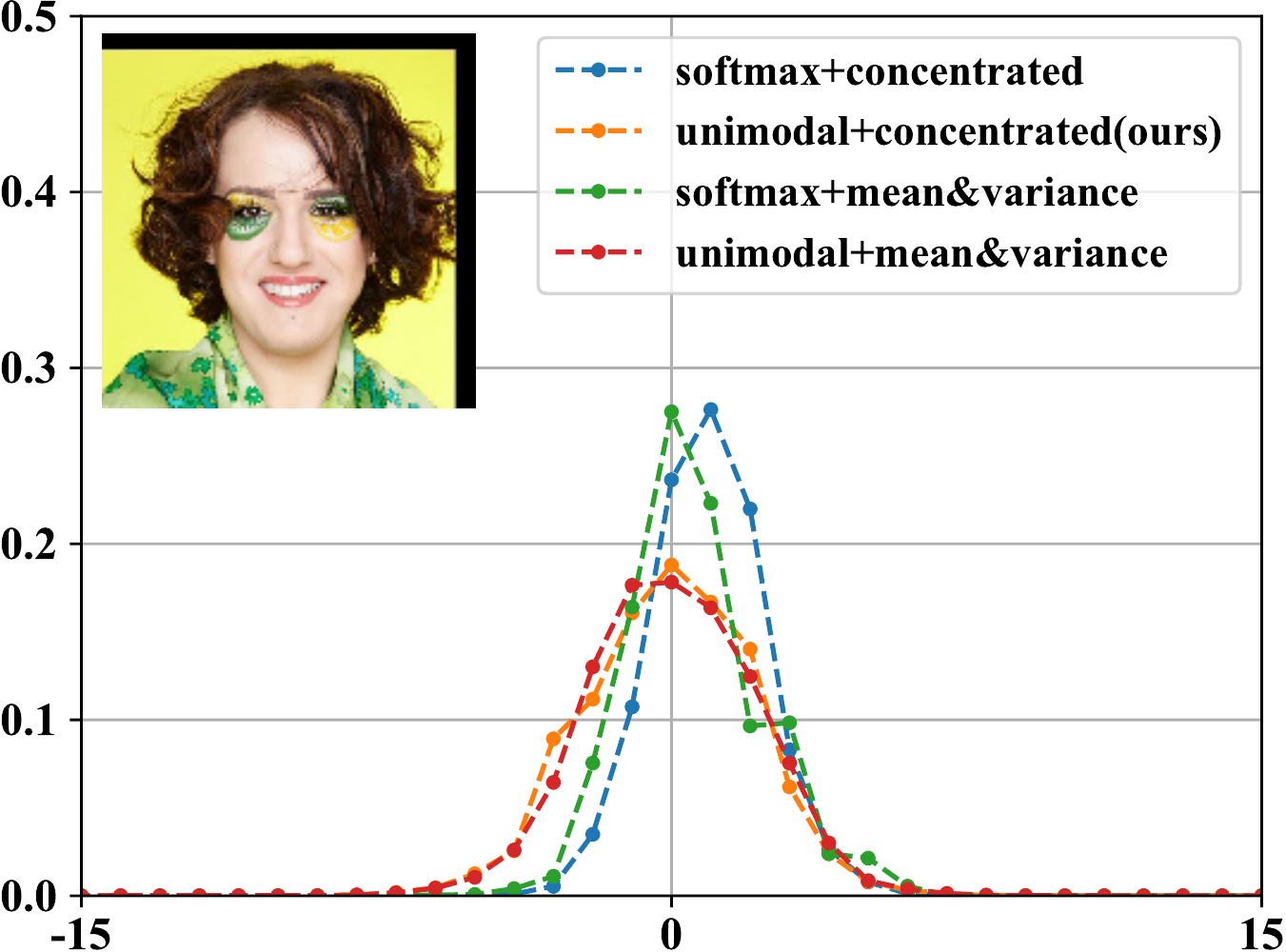} \quad
	\includegraphics[width=0.2 \linewidth]{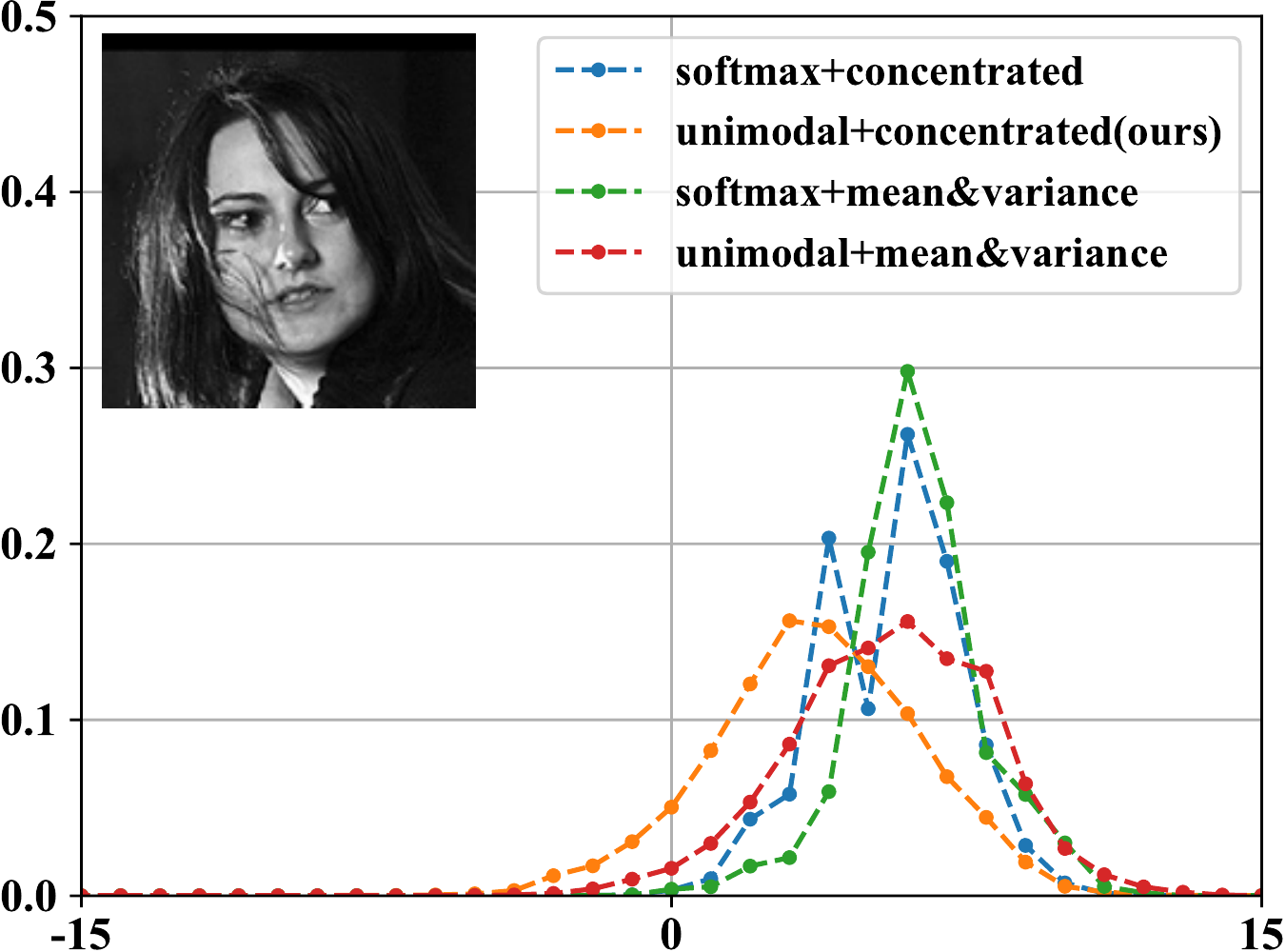}
	\caption{Examples of head pose estimation for different losses combinations when the yaw angle is zero. The first row examples are in high quality
and the second row examples are in low quality which are polluted by illumination, occlusion and heavy makeup. As can be see, first, when equipped with unimodal loss, the distributions are relatively smooth and unimodal. When equipped with softmax loss, the distributions are easy to be multimodal. Second, when equipped with concentrated loss, the concentrations of distribution or the prediction uncertainties vary obviously among the samples of the same class with different quality. }
	\label{fig:ages_headpose_std}
\end{figure*}

\begin{figure}[t]
	\begin{center}
		\includegraphics[width=0.65\linewidth]{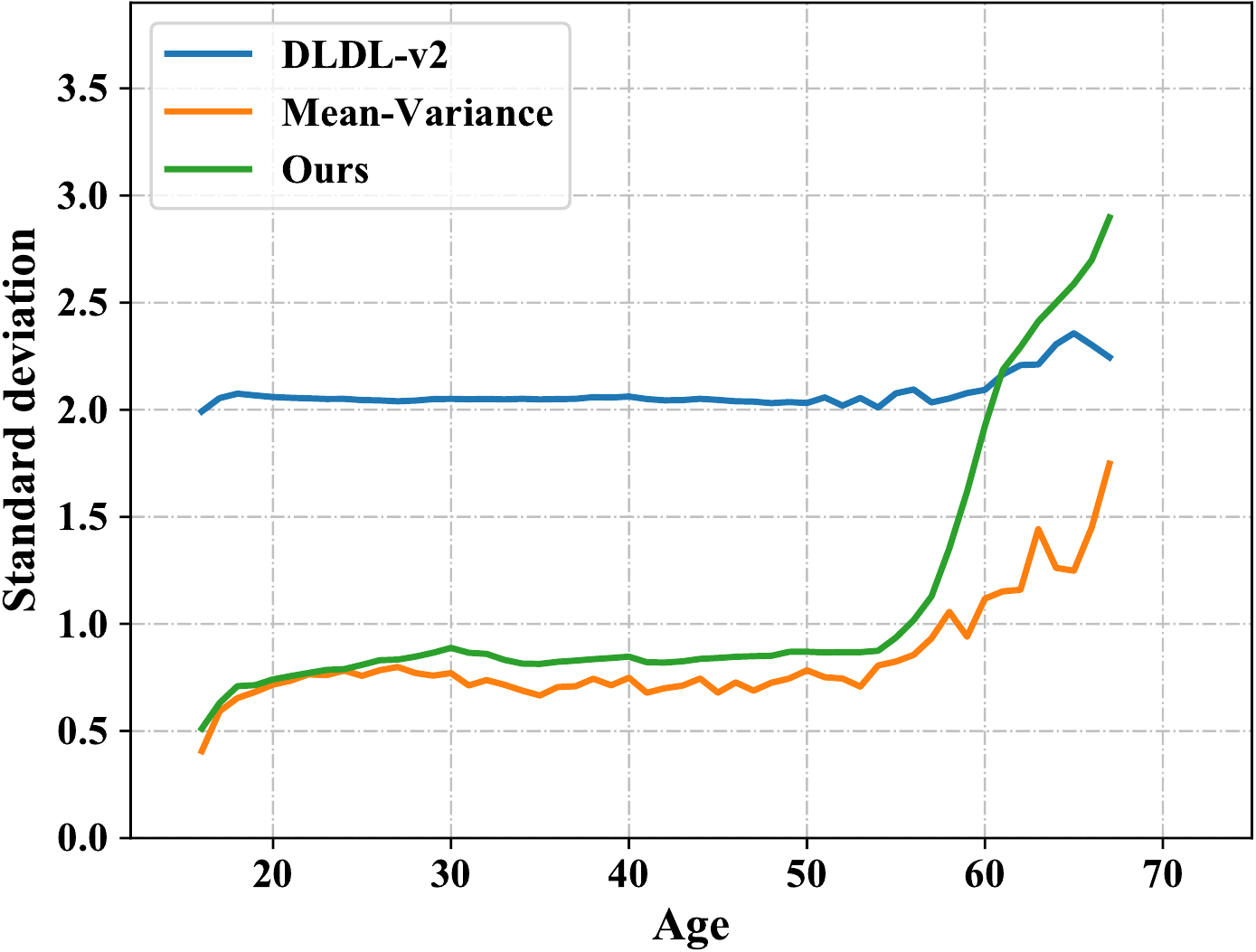}
	\end{center}
	\caption{ Average standard deviations at different labels on MORPH II. Labels with the number of samples less than 10 are discarded.}
	\label{fig:visualization_1}
\end{figure}


\subsection{Comparison with the State-of-the-arts}
In this section, we compare our methods with state-of-the-art ones on Morph II, AFLW2000 and BIWI respectively.
As shown in Table \ref{table:age_compare}, our model for age estimation achieves 1.86 MAE with the VGG-16 backbone, which is the best performance among all methods. It is noted that compared with the FLDL based SPUDFRs \cite{2020SPDFR} and ALDL based Mean-Variance \cite{2018Mean}, our result is obviously better than them which shows the effectiveness of our proposed fully adaptive label distribution learning.

As shown in Table \ref{tabel:compare_head_pose}, on the challenging AFLW2000 and BIWI datasets for head pose estimation, our unimodal-concentrated loss outperforms previous state-of-the-art methods such as FSA \cite{2020FSA} and FDN \cite{2020FDN}, which further exhibits its superiority. Moreover, compared
with landmark-based methods \cite{2016Face, 2017How}, our method only uses
pixel intensity information which is landmark-free.

\textbf{Comparison with different losses}.
Methods for ordinal regression listed in Table \ref{table:age_compare} and Table \ref{tabel:compare_head_pose} are all under different experimental settings.
For fair comparison, we conduct the experiment using the VGG-16 backbone with different losses. Specifically, we choose DLDL-v2 and Mean-Variance as they achieve the convincing performances based on FLDL and ALDL respectively. As shown in Table \ref{table:compare_fixed_adaptive_distribution}, our proposed loss outperforms the DLDL-v2 and Mean-Variance loss.
As viewed in Fig. \ref{fig:visualization_1}, DLDL-v2 loss tends to output distributions with the same variance at different ages, because it learns from a fixed-form distribution with the assumed standard deviation. Mean-Variance loss generates distributions with smaller variances than ours as the distributions are optimized to be as sharp as possible. While the learned distributions of our unimodal-concentrated loss can adapt more appropriately with the facial aging. Some examples are shown in Fig. \ref{fig:age_figure_new}.

\subsection{Ablation Study}

\subsubsection{Different Loss Combinations}
Our proposed loss is composed of the unimodal loss and concentrated loss.
Since it is hard to optimize the network with only one part, to demonstrate the effectiveness of each part respectively, we conduct experiments under different combinations of our loss with loss in Mean-Variance. From Table \ref{tabel:cross_validation}, we can see:

Compared with the combination of softmax and concentrated loss, the combination of unimodal and concentrated loss achieves higher performance, since the unimodal loss constrains the probabilities to be unimodal while the softmax loss not. As shown in the second image and the last image of the second row of Fig \ref{fig:ages_headpose_std}, the probabilities outputed by softmax+concentrated is multimodal. It verifies the effectiveness of our proposed principle \uppercase\expandafter{\romannumeral2}, i.e. the distribution should be unimodal.

Compared with the combination of softmax and mean loss \& variance loss, the combination of softmax and concentrated loss achieves higher performance since concentrated loss takes instance-aware uncertainty into consideration instead of minimizing the mean and variance loss as small as possible in \cite{2018Mean}. As shown in Fig \ref{fig:ages_headpose_std}, the confidences outputed by softmax+concentrated in the normal faces (shown in the first row) is relatively higher, and the ones in the hard faces (shown in the second row) is relatively lower. While the confidences output by softmax+mean\&variance have relatively smaller difference between the two rows. It verifies the effectiveness of our proposed principle \uppercase\expandafter{\romannumeral3}, i.e. the label distribution should vary with the samples changing.

It is worth noting that, compared with the combination of softmax and mean loss \& variance loss, although the combination of unimodal and mean loss \& variance loss can make the probabilities to be unimodal as shown in Fig \ref{fig:ages_headpose_std}, it still gets the poorer performance. The reason is that it is hard to optimize the network with the mean \& variance loss when the softmax loss is not used jointly as viewed in \cite{2018Mean}, while our concentrated loss does not have this problem.
More detailed analysis can be found in the supplementary materials.

\section{Conclusion}
In this paper, we propose a fully adaptive distribution learning method for ordinal regression by introducing an efficient cost function called unimodal-concentrated loss. The unimodal loss ensures the unimodality of the learned distribution and the concentrated loss maximizes the probability at the ground-truth in a fully adaptive way for individual instances. Experimental results show our method outperforms previous works on MORPH II benchmark for facial age estimation, AFLW2000 and BIWI benchmarks for head pose estimation. In the future work, we would like to investigate the effectiveness of the proposed loss in other related tasks.

{\small
\bibliographystyle{ieee_fullname}
\bibliography{egbib_cvpr2022}
}

\clearpage
\onecolumn

\section*{\uppercase\expandafter{\romannumeral1} The influences of hyper-parameter $\lambda$  for our loss}
As shown in Tab.\ref{table:ablation_of_hyper_lambda}, the network is difficult to converge if $\lambda$ is too small (concentrated loss takes a dominant role) and there will be a big performance degradation if $\lambda$ is too large (the unimodal loss takes a dominant role). Within a long reasonable range, our method performs stably.

\begin{table}[h]
		\caption{The influences of hyper-parameter $\lambda$  for our loss. }
		\begin{center}
			\begin{tabular}{|p{1.8cm}<{\centering}|p{0.8cm}<{\centering}|p{0.8cm}<{\centering} |p{0.8cm}<{\centering} |p{0.8cm}<{\centering}| p{0.8cm}<{\centering} |p{0.8cm}<{\centering}|  }
				\hline
				Threshold $\lambda$ & 1e-1 & 1e1 & 1e2 & 1e3 & 2e3 & 1e4 \\
				\hline
				MORPH II & NaN& 1.92 & 1.88 & 1.86 & 1.88 & 3.24 \\ \hline
				AFLW2000 & NaN& 4.21 & 4.26 & 4.13 & 4.11 & 6.12 \\ \hline
				BIWI & NaN& 3.67 & 3.71 & 3.57 & 3.61 & 5.75 \\ \hline
			\end{tabular}
		\end{center}
		\label{table:ablation_of_hyper_lambda}
\end{table}

\section*{\uppercase\expandafter{\romannumeral2} Demonstration for softmax+mean \& variance loss superior to unimodal+mean \& variance loss}
The Mean-Variance loss \cite{2018Mean} can be formulated as
\begin{align}
	L_{m-v}&=L_{s} + \lambda_{1}L_{m} + \lambda_{2}L_{v} \\
   	  &=\frac{1}{N}\sum_{i=1}^{N}{-}logp_{i,y_{i}} +\frac{\lambda_{1}}{2}(\hat{y_{i}} - y_{i})^2 + \lambda_{2}v_{i},
	\label{eq:mean-variance}
\end{align}
where $L_{s}$ is the softmax loss.

We first demonstrate that it is hard for the network to directly optimize mean loss and variance loss without softmax loss. Based on Eq. \ref{eq:pik} and derivation process in \cite{2018Mean}, the gradient of $L_{m}$ w.r.t. $z_{i,j}$ can be computed as
\begin{equation}
\frac{\partial L_{m}}{\partial z_{i,j}} = \frac{\hat{y_{i}}-y_{i}}{N} p_{i,j}(j-\hat{y_{i}}).
\label{eq:gradient_supple_1}
\end{equation}

According to the Eq. \ref{eq:gradient_supple_1} , as analyzed in \cite{2018Mean}, for an estimated distribution with mean value $\hat{y_{i}}$, if $\hat{y_{i}} < y_{i}$, the network will be updated to increase the probabilities of the classes $j (j > \hat{y_{i}})$ via their negative gradients, and decrease the probability of those classes $j (j < \hat{y_{i}})$ via their positive gradients. In this way, the mean value of the estimated distribution will be increased,
and becomes closer to $y_{i}$.

The gradient of $L_{v}$ w.r.t. $z_{i,j}$ can be computed as
\begin{equation}
\frac{\partial L_{v}}{\partial z_{i,j}} = \frac{1}{N} p_{i,j}((j-\hat{y_{i}})^2 - v_{i}).
\label{eq:gradient_supple_2}
\end{equation}

The gradient in Eq. \ref{eq:gradient_supple_2} has the following properties:
\begin{equation}
j \in (\hat{y_{i}}-\sqrt{v_{i}}, \hat{y_{i}}+\sqrt{v_{i}}), \frac{\partial L_{v}}{\partial z_{i,j}} < 0,
\label{eq:gradient_supple_3}
\end{equation}
and
\begin{equation}
j \in [1, \hat{y_{i}}-\sqrt{v_{i}}) \cup (\hat{y_{i}}+\sqrt{v_{i}}, C ], \frac{\partial L_{v}}{\partial z_{i,j}} > 0.
\label{eq:gradient_supple_4}
\end{equation}

As analyzed in \cite{2018Mean}, Eq. \ref{eq:gradient_supple_3} shows that, the network will be updated to increase
the probabilities of the classes $j$ close to $\hat{y_{i}} (j \in (\hat{y_{i}}-\sqrt{v_{i}}, \hat{y_{i}}+\sqrt{v_{i}}))$ via their negative gradients. On the
contrary, Eq. \ref{eq:gradient_supple_4} shows that the network will be updated to decrease the probabilities of the classes $j$ far away from $\hat{y_{i}} (j \in [1, \hat{y_{i}}-\sqrt{v_{i}}) \cup (m_{i}+\sqrt{v_{i}}, C ])$  via their positive gradients.

Base on analysis above, it can be observed that
\begin{equation}
\begin{aligned}
&\text{if} \quad j \in (\hat{y_{i}}-\sqrt{v_{i}}, \hat{y_{i}}) \quad  \text{and} \quad  \hat{y_{i}} < y_{i} \\& \text{then} \quad \frac{\partial L_{m}}{\partial z_{i,j}} > 0 \quad,  \quad \frac{\partial L_{v}}{\partial z_{i,j}} < 0.
\label{eq:gradient_supple_5}
\end{aligned}
\end{equation}

In the case of Eq. \ref{eq:gradient_supple_5}, let $ \mid \frac{\partial L_{v}}{\partial z_{i,j}} \mid \quad > \quad \mid \frac{\partial L_{m}}{\partial z_{i,j}} \mid $ ($\lambda_{1}$ and $\lambda_{2}$ are omitted for demonstration)
\begin{equation}
\begin{aligned}
& \Rightarrow v_{i} - (j-\hat{y_{i}})^2 > (\hat{y_{i}} - y_{i})(j - \hat{y_{i}}), \\& \Rightarrow v_{i} > (j-y_{i})(j-\hat{y_{i}}).
\label{eq:gradient_supple_6}
\end{aligned}
\end{equation}

\begin{figure}[h]
	\begin{center}
		\includegraphics[width=0.8\linewidth]{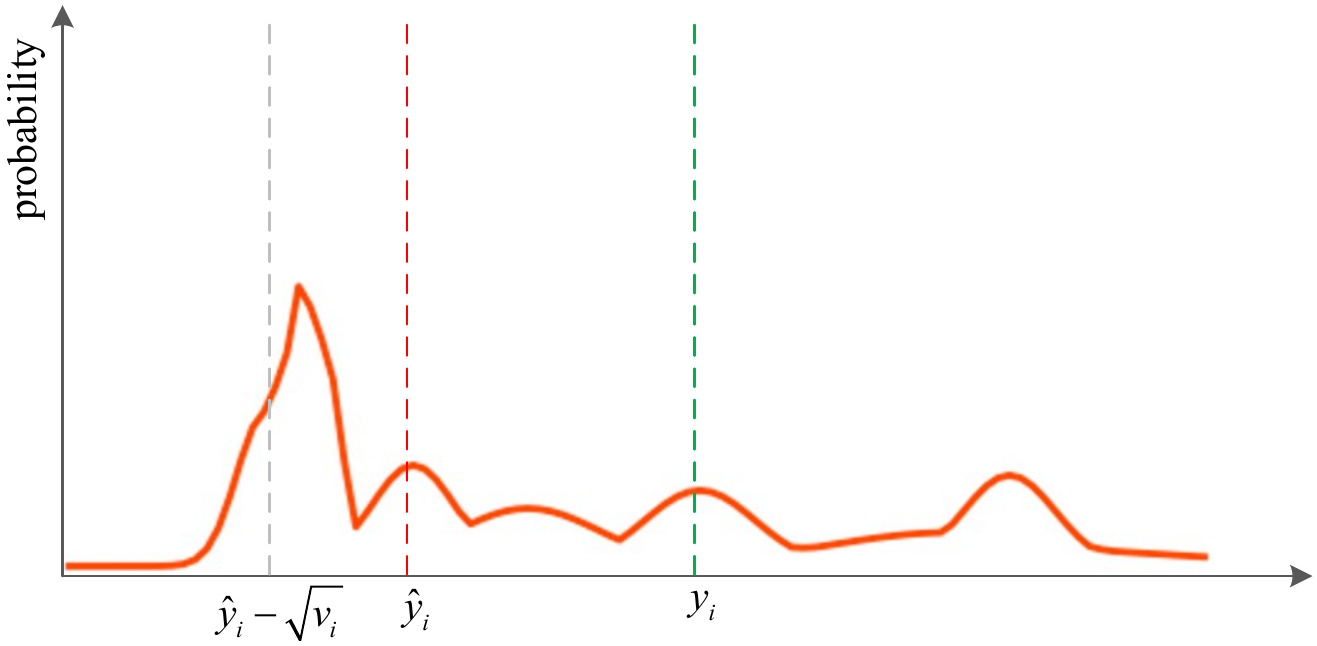}
	\end{center}
	\caption{A typical example of distribution at the early stage of training.
	}
	\label{fig:curve_sup}
\end{figure}

According to the Eq. \ref{eq:gradient_supple_6}, when $ v_{i} > (j-y_{i})(j-\hat{y_{i}})$, the absolute value of $\frac{\partial L_{v}}{\partial z_{i,j}} $ is larger than that of $\frac{\partial L_{m}}{\partial z_{i,j}} $. Consequently, the network will be updated to increase the probabilities of the classes $j (j \in (\hat{y_{i}}-\sqrt{v_{i}}, \hat{y_{i}})$ which are far from the ground-truth $ y_{i} $. That is to say, when large fluctuation appears at the early stage of training \cite{2018Mean} which meets the such condition, the probabilities of the classes far from the ground-truth $ y_{i} $ will be increased and it is against principle \uppercase\expandafter{\romannumeral1}. It accounts for that it is hard to optimize the network with the mean \& variance loss only. A typical example corresponding to this condition is given in Fig. \ref{fig:curve_sup}.

When adding the softmax loss, as we all know, the gradient of $L_{s}$ w.r.t. $z_{i,j}$ can be computed as
\begin{align}
\frac{\partial L_{s}}{\partial z_{i,j}}
=
\begin{array}{lr}
p_{i,j}- y_{i,j} \\
\end{array}
\label{eq:gradient_supple_7}
\end{align}
where $y_{i,j}$ is the indicator whether the instance $i$ belongs to class $j$. If instance $i$ belongs to class $j$, $y_{i,j}=1$, otherwise, $y_{i,j}=0$.
According to the Eq. \ref{eq:gradient_supple_7}, it can be seen that the network will always be updated to increase the probability of the class $y_{i}$ via their negative gradients. It accounts for that softmax loss can promote the network to converge with the mean \& variance loss.

When adding the unimodal loss, the gradient of $L_{uni}$ w.r.t. $z_{i,j}$ can be computed as
\begin{align}
\frac{\partial L_{uni}}{\partial z_{i,j}}=\frac{\partial L_{uni}}{\partial p_{i,j}} \frac{\partial p_{i,j}}{\partial z_{i,j}}=
\left\{
\begin{array}{lr}
p_{i,j}(1 - p_{i,j}),  \quad \  (p_{i,j}{-}p_{i,j+1})*\text{sign}[j-y_{i}] < 0 \\
0,  \quad \quad \quad \quad \quad \quad (p_{i,j}{-}p_{i,j+1})*\text{sign}[j-y_{i}] >= 0, \\
\end{array}
\right.
\label{eq:gradient_supple_8}
\end{align}

the gradient of $L_{uni}$ w.r.t. $z_{i,j+1}$ can be computed as
\begin{align}
\frac{\partial L_{uni}}{\partial z_{i,j+1}}=\frac{\partial L_{uni}}{\partial p_{i,j+1}} \frac{\partial p_{i,j+1}}{\partial z_{i,j+1}}=
\left\{
\begin{array}{lr}
-p_{i,j+1}(1 - p_{i,j+1}),  \quad \ \ (p_{i,j}{-}p_{i,j+1})*\text{sign}[j-y_{i}] < 0 \\
0,  \quad \quad \quad \quad \quad \quad \quad \quad \quad (p_{i,j}{-}p_{i,j+1})*\text{sign}[j-y_{i}] >= 0. \\
\end{array}
\right.
\label{eq:gradient_supple_9}
\end{align}
According to the Eq. \ref{eq:gradient_supple_8} and \ref{eq:gradient_supple_9}, it can be seen that
our unimodal loss aims at correcting the ordinal relationship when two neighboring probabilities are ranked by mistake instead of directly maxmizing the probability of the class $y_{i}$ like softmax loss. So, compared with the combination of softmax and mean loss \& variance loss, the combination of unimodal and mean loss \& variance loss gets the poorer performance.

\section*{\uppercase\expandafter{\romannumeral3} Ablation study about single unimodal loss and single concentrated loss.}
\begin{figure}[h]
	\begin{center}
		\includegraphics[width=0.8\linewidth,height=0.3\linewidth]{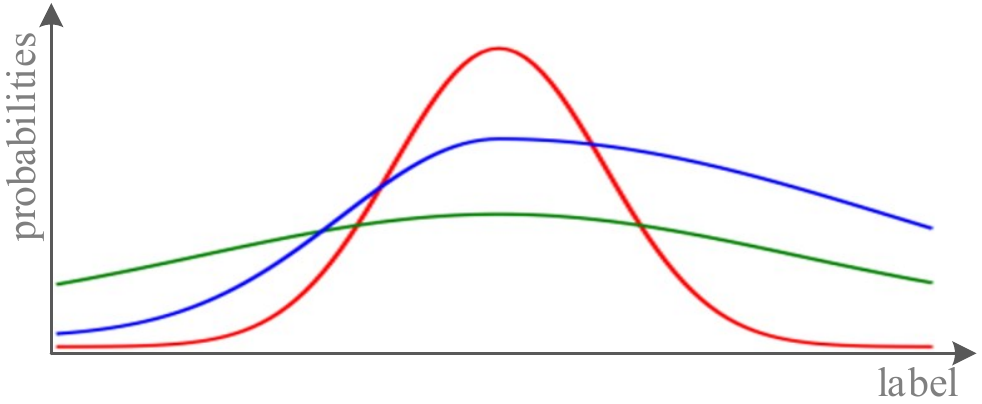}
	\end{center}
	\caption{Typical distribution examples. }
	\label{fig:distribution_examples}
\end{figure}

Without the unimodal loss, the network is difficult to converge. The MeanVariance loss encounters the similar problem without softmax loss as mentioned by the paper \cite{2018Mean} in Sec. 3.2. Therefore, in ablation as shown in Tab. \ref{tabel:cross_validation} of main script, only our concentrated loss along with softmax loss or unimodal loss and MeanVariance loss along with softmax loss or unimodal loss are campared. Without the concentrated loss, although the network can converge. However, with only the unimodal loss, different distributions (shown as the three samples in Fig. \ref{fig:distribution_examples}) may have the same loss value, but very distinct shapes, which may make the network converge to a bad local minimum. Its age estimation error is larger than 5 on MORPH II, which verifies this.

\begin{table}[h]
	\caption{The results for the single unimodal loss and single concentrated loss.}
	\begin{center}
		\small
		\setlength{\tabcolsep}{1.2mm}{
			\begin{tabular}{|c|c|c|c|c|c|c|}
				\hline
				\multicolumn{2}{|c|}{Combinations} & \multicolumn{3}{c|}{Benchmarks} \cr \cline{1-5}
				Auxiliary& Primary & MORPH II & AFLW2000 & BIWI \\ \hline \hline
				Softmax &Concentrated& 1.92 & 4.25 & 3.61 \\ \hline
				Unimodal &Concentrated&1.86 & 4.13 & 3.57 \\ \hline
				Softmax &Mean \& Variance &2.01 & 4.36 & 4.01 \\ \hline
				Unimodal &Mean \& Variance & 3.30 & 4.53 & 4.39 \\ \hline
				Unimodal & - & $\ge$ 5 & $\ge$ 7 & $\ge$ 7 \\ \hline
				- &Concentrated& NaN & NaN & NaN \\ \hline
			\end{tabular}
		}
	\end{center}
	\label{tabel:cross_validation}
\end{table}

\end{document}